\documentclass[10pt,twocolumn,letterpaper]{article}

\usepackage{cvpr}
\usepackage{times}
\usepackage{epsfig}
\usepackage{graphicx}
\usepackage{amsmath}
\usepackage{bm}
\usepackage{amssymb}

\newcommand{\ve}{\mathbf{e}}

\newcommand{\vpi}{\bm{\pi}}

\newcommand{\semean}{\bm{\mu}}
\newcommand{\sevar}{\bm{\phi}}

\newcommand{\vZ}{\mathbf{Z}}
\newcommand{\vS}{\mathbf{S}}
\newcommand{\vT}{\mathcal{T}}
\newcommand{\vN}{\mathcal{N}}

\cvprfinalcopy 

\def\cvprPaperID{****} 

\ifcvprfinal\fi
\begin{document}

\title{Epitome for Automatic Image Colorization}

\author{Yingzhen Yang${}^{1}$ Xinqi Chu${}^{1}$ Tian-Tsong Ng${}^{2}$ Alex Yong-Sang Chia${}^{2}$ Shuicheng Yan${}^{3}$ Thomas S. Huang${}^{1}$\\
\small ${}^{1}$Department of Electrical and Computer Engineering, University of Illinois at Urbana-Champaign\\
\small${}^{2}$Institute for Infocomm Research, Singapore\\
\small${}^{3}$Department of Electrical and Computer Engineering, National University of Singapore, Singapore\\
\small$\left\{yyang58, chu36, huang\right\}@ifp.uiuc.edu$, $\left\{ttng, ysachia\right\}@i2r.a$-$star.edu.sg$, $eleyans@nus.edu.sg$
}

\maketitle

\thispagestyle{empty}


\begin{abstract}
   Image colorization adds color to grayscale images. It not only increases the visual appeal of grayscale images, but also enriches the information contained in scientific images that lack color information. Most existing methods of colorization require laborious user interaction for scribbles or image segmentation. To eliminate the need for human labor, we develop an automatic image colorization method using epitome. Built upon a generative graphical model, epitome is a condensed image appearance and shape model which also proves to be an effective summary of color information for the colorization task. We train the epitome from the reference images and perform inference in the epitome to colorize grayscale images, rendering better colorization results than \cite{WelshAM02} in our experiments.
\end{abstract}

\section{Introduction}
Colorization adds color to grayscale images by assigning color values to images which only contain a grayscale channel. It not only increases the visual appeal, but also enhances the information conveyed by scientific images. For example, the grayscale images  acquired by scanning electron microscopy (SEM) can be made more illustrative by adding different colors to different parts of the images. However, the manual colorization is tedious and time consuming, so it is not suitable for batch process. To overcome this problem, we propose an automatic colorization method by epitome. Figure~\ref{fig:nano} shows the colorization result for the Nano Mushroom-like image. We train the epitome from one manually colorized Nano Mushroom-like image, and use that epitome to automatically colorize the other Nano Mushroom-like image, which eliminates the need for human labor and makes batch colorization process possible.

Based on the source of the color information used to colorize the grayscale images, existing colorization techniques fall into two main categories: user scribble based methods and color transfer methods. The user scribble based method in \cite{LevinLW04} asked users to draw color scribbles in the grayscale image, and the algorithm propagated the user-provided color to the whole image requiring that similar neighboring pixels should receive similar color. Later, L. Qing et al. \cite{LuanWCLXS07} proposed a method which required less human intervention. The user scribbles were employed for texture segmentation and user-provided color was propagated within each segment. Using a similar color image as a reference, the color transfer methods such as \cite{WelshAM02} performed colorization by transferring the color from the reference image to the grayscale image, either automatically or with user intervention. However, the pixel-level matching based on luminance value and neighborhood statistics adopted by \cite{WelshAM02} suffered from spatial inconsistency and the user-provided swatches were required to guide the matching process in many cases. \cite{IronyCL05} improved the spatial consistency by an image space voting scheme. Their method first transferred color to a few pixels in the target image with high confidence, then applied the method in \cite{LevinLW04} to colorize the whole image, treating the colorized pixels in the first step as the scribbles. However, their method required a robust segmentation of the reference image, which was difficult in many cases without user intervention.

Similar to \cite{WelshAM02}, our automatic colorization method transfers the color information from the reference image to the target grayscale image.
Since most of existing colorization methods need user interactions for color selection or segmentation, a robust and automatic colorization algorithm is preferable. In order to approach this problem, it is worthwhile to exploit the biological characteristics of human visual system. The average human retina contains much more rods than cones \cite{CurcioEtAl1990} (92 million rods versus 4.6 million cones). Rods are more sensitive to cones but they are not sensitive to color, so that most of visually significant variation arises only from luminance differences. This fact suggests that we do not need to search the whole reference image for the color patches to colorize the target image, instead we can reduce the search space for color patches, or equivalently find an effective color summary of the reference image, to improve the efficiency and alleviate color assignment ambiguity. In \cite{WelshAM02}, such summary is a set of source color pixels randomly sampled, which is, however, subject to noise in the raw pixels.

In order to find an effective and compact summary of the color information in the reference image, we adopt the condensed image appearance and shape representation, i.e. epitome \cite{JojicFK03}. Epitome consolidates self-similar patches in the spatial domain, and the size of the epitome is much smaller than that of the image it models. By virtual of the generative graphical model, epitome can be interpreted as a tradeoff between template and histogram for image representation and it has been applied to many computer vision tasks such as object detection, location recognition and synthesis \cite{Ni09, ChuYLCH10}. Epitome summarizes a large number of raw patches in the reference image by only representing the most constitutive elements. In our epitomic colorization scheme the color patches used to colorize the target grayscale image are retrieved from the epitome trained with the reference image, rather than from the raw image patches. Epitome proves to an effective summary of the color information in the reference image, which produces more satisfactory colorization results than \cite{WelshAM02} in the experiments.

The paper is arranged as follows. Section 2 describes the process of automatic colorization by epitome as well as the detailed formulation of training the epitome and inference in the epitome graphical model, especially on how epitome summarizes the raw image patches of the reference image into a condensed representation and how inference is performed in epitome to automatically colorize the target grayscale image. Section 3 shows the colorization results, and we conclude the paper in Section 4.

\section{Formulation}

\subsection{Description of Automatic Colorization by Epitome}
Given a reference color image $cI$ and the target grayscale image $gI$, we aim to automatically colorize $gI$ with the color information from $cI$. We achieve this goal by first training an epitome $e$ from the reference image, then performing inference in $e$ so as to transfer the color information of the color patches of $\hat{\ve}$ to the corresponding grayscale patches of $gI$. Note that the grayscale channel of $gI$ is retained as the luminance channel after the color transfer process. We will illustrate the training and inference process in detail in the following subsections.

\subsection{Training the Epitome}
Epitome is a latent representation of an image, which comprises hidden variables and parameters required to generate the image patches according to the epitome graphical model. Epitome summarizes a large set of raw image patches into a condensed representation of a size much smaller than the original image, and it approaches this goal in a manner similar to Gaussian Mixture Model with overlapping means and variances.

The epitome $e$ of an image $I$ of size $M \times N$ is a condensed representation of size ${M_e} \times {N_e}$ where ${M_e} < M$ and ${N_e} < N$. The epitome contains two parameters: $\ve = \left( {\semean ,\sevar} \right)$. $\semean$ and $\sevar$ represent the Gaussian mean and variance respectively and both of them are of size ${M_e} \times {N_e}$. Suppose $Q$ patches are sampled from the reference image, i.e. $\{\vZ_k\}^Q_{k=1}$, and each patch $\vZ_k$ contains pixels with image coordinates $\vS_k$. Similar to \cite{JojicFK03}, the patches are square and we use fixed patch size throughout this paper. These patches are densely sampled and they can be overlapping with each other to cover the entire image. We associate each patch $\vZ_k$ with a hidden mapping $\vT_k$ which maps the image coordinates $\vS_k$ to the epitome coordinates, and all the $Q$ patches are generated independently from the epitome parameters and the corresponding hidden mappings as below:

\begin{equation} \label{eq:patchfromT}
p(\vZ_k | \vT_k, \ve) =  \prod_{i \in \vS_k}  \vN(z_{i,k} ; \semean_{\vT_k(i)}, \sevar_{\vT_k(i)}), k = 1..Q
\end{equation}

and
\begin{equation} \label{eq:patchfromT1}
\prod_{k=1}^Q p(\{\vZ_k\}^Q_{k=1} | \{\vT_k\}^Q_{k=1}, \ve) = \prod_{k=1}^Q p(\vZ_k | \vT_k, \ve)
\end{equation}

where $z_{i,k}$ is the pixel with image coordinates $i$ from the $k$-th patch. Since $z_{i,k}$ is independent of the patch number $k$, we simply denote it as $z_i$ in the following text. $\vN(\cdot;\mu,\phi)$ represents a Gaussian distribution with mean $\hat\mu$ and variance $\hat\phi$
$$\vN(\cdot; \hat{\mu},\hat{\phi})  = \frac{1}{\sqrt{2 \pi \hat{\phi}}} \exp^{-\frac{(\cdot-\hat{\mu})^2}{2\hat{\phi}}}.$$

\begin{figure}[t]
\begin{center}
\includegraphics[scale=.45]{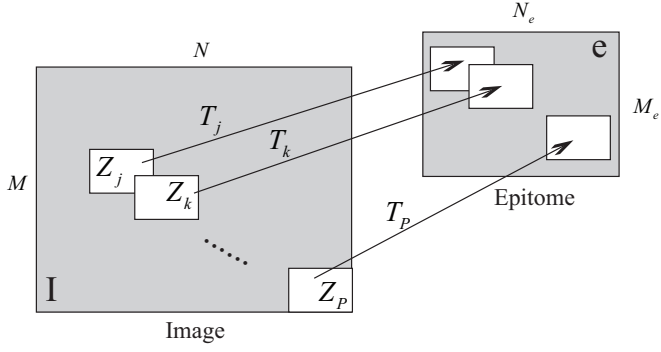}
\end{center}
   \caption{The mapping $\vT_k$ maps the image patch $\vZ_k$ to its corresponding epitome patch with the same size, and $\vZ_k$ can be mapped to any possible epitome patch according to $\vT_k$.}
\label{fig:hiddenmapping}
\end{figure}

\begin{figure}[t]
\begin{center}
\includegraphics[scale=.5]{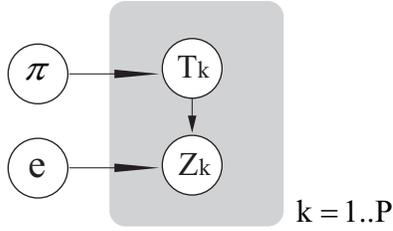}
\end{center}
   \caption{The epitome graphical model}
\label{fig:epitomegm}
\end{figure}

Based on (\ref{eq:patchfromT}), the hidden mapping $\vT_k$ can be interpreted as a hidden variable that indicates the location of the epitome patch from which the observed image patch $\vZ_k$ is generated, and it behaves similar to the hidden variable in the traditional Gaussian mixture models that specifies the Gaussian component from which a specific data point is generated. Also, $\vT_k$ maps the image patch to its corresponding epitome patch, and the number of possible mappings that each $\vT_k$ can take, denoted as $L$, is determined by all the discrete locations in the epitome ($L = {M_e} \times {N_e}$ in our setting).  Figure~\ref{fig:hiddenmapping} illustrates the role that the hidden mapping variables play in the generative model, and Figure~\ref{fig:epitomegm} shows the epitome graphical model, which again demonstrate its similarity to Gaussian mixture models. $\vpi  \buildrel \Delta \over = \left\{ {{\pi _l}} \right\}_{l = 1}^L$ indicates the prior distribution of the hidden mapping. Suppose $\vT_{k,l}$ is the $l$-th mapping that $\vT_k$ can take, then

$$ p(\vT_k) = \prod_{l=1}^L {\pi_l}^{\delta \left( {\vT_k=\vT_{k,l}} \right)}$$

\noindent which holds for any $k \in \left\{ {1..Q} \right\}$. $\delta$ is an indicator function and $\delta$ equals to $1$ when its argument is true, and $0$ otherwise.

Our goal is to find the epitome $\hat{\ve}$ that maximizes the log likelihood function:

\begin{equation} \label{eq:eloglikelihood}
{\hat{\ve}} = \mathop {\arg \max }\limits_\ve \log p\left( {\{\vZ_k\}^Q_{k=1}|\ve} \right)
\end{equation}

Given the epitome $\ve$, the likelihood function for the complete data, i.e. the image patches $\{\vZ_k\}^Q_{k=1}$ and the hidden mappings $\{\vZ_k\}^Q_{k=1}$, is derived below according to the epitome graphical model:

\begin{eqnarray}
&&p(\{\vZ_k, \vT_k\}^Q_{k=1} |\ve,\vpi) = \prod_{k=1}^Q p( \vZ_k, \vT_k|\ve,\vpi) \nonumber \\
&&=\prod_{k=1}^Q p(\vT_k) p( \vZ_k | \vT_k,\ve) \nonumber \\
&&= \prod_{k=1}^Q \prod_{l=1}^L \left[ {{\pi_l} \prod_{j \in \vS_k} \vN({z_j} ; \semean_{\vT_{k,l}(j)}, \sevar_{\vT_{k,l}(j)})} \right]^{\delta \left( {\vT_k=\vT_{k,l}} \right)}
\end{eqnarray}

We use the Expectation-Maximization algorithm \cite{Dempster77} to maximize the likelihood function (\ref{eq:eloglikelihood}) and learn the epitome $\hat{\ve}$, following the procedure introduced in \cite{Bishop06}.

\textbf{The E-step:}
The posterior distribution of the hidden variables, i.e. the hidden mapping is

\begin{align} \label{eq:estep}
&q(\vT_k)  \buildrel \Delta \over =  p(\vT_k |\vZ_k,\ve,\vpi) \nonumber \\
&=\frac{ p(\vZ_k | \vT_k,\ve) p(\vT_k)} {\sum_{\vT_k} p(\vZ_k | \vT_k,\ve) p(\vT_k)} \nonumber \\
&=\frac{ \prod_{l=1}^L \left[ {{\pi_l} \prod_{j \in \vS_k} \vN({z_j} ; \semean_{\vT_{k,l}(j)}, \sevar_{\vT_{k,l}(j)})} \right]^{\delta \left( {\vT_k=\vT_{k,l}} \right)} } {\sum_{\vT_k} \prod_{l=1}^L \left[ {{\pi_l} \prod_{j \in \vS_k} \vN({z_j} ; \semean_{\vT_{k,l}(j)}, \sevar_{\vT_{k,l}(j)})} \right]^{\delta \left( {\vT_k=\vT_{k,l}} \right)}} \nonumber \\
\end{align}

We observe that $q(\vT_k)$ corresponds to the responsibility in Gaussian mixture models.

\textbf{The M-step:}
We obtain the expectation of the log-likelihood function for the complete data with respect to the posterior distribution of the hidden mapping from the E-step as below:

\begin{align} \label{eq:logcompletedata}
&E\left[ {\log p\left( {\{\vZ_k, \vT_k\}^Q_{k=1}|\ve,\vpi } \right)} \right] \nonumber \\
&=\sum\limits_{k = 1}^Q {\sum\limits_{l = 1}^L {q(\vT_k = \vT_{k,l}) \cdot \left[ {\log {\pi _l} + \log p\left( {{\vZ_k}|\vT_k = \vT_{k,l},\ve} \right)} \right]} }
\end{align}

Maximizing (\ref{eq:logcompletedata}) with respect to $\left( {\ve,\vpi } \right)$, we get the following update of the parameters of the epitome and $\vpi$:

\begin{equation} \label{eq:Mstepemean}
{\semean}_j = \frac{\sum\limits_{k = 1}^Q \sum_{i \in \vS_k} \sum_{\vT_k} \delta(\vT_k(i) = j)q(\vT_k)z_i}{\sum\limits_{k = 1}^Q \sum_{i \in \vS_k} \sum_{\vT_k} \delta(\vT_k(i) = j)q(\vT_k) }
\end{equation}

\begin{equation} \label{eq:Mstepevar}
{\sevar}_j = \frac{\sum\limits_{k = 1}^Q \sum_{i \in \vS_k} \sum_{\vT_k} \delta(\vT_k(i) = j)q(\vT_k) (z_i - {\semean}_j)^2}{\sum\limits_{k = 1}^Q \sum_{i \in \vS_k} \sum_{\vT_k} \delta(\vT_k(i) = j)q(\vT_k) }
\end{equation}

\begin{equation} \label{eq:Mstepepi}
{\pi _l} = \frac{{\sum\limits_{k = 1}^Q {p\left( {{\vT_k} = {\vT_{k,l}}} \right)} }}{Q}, l=1..L
\end{equation}

The index $j$ indicates the epitome coordinates in (\ref{eq:Mstepemean}) and (\ref{eq:Mstepevar}). We alternate between E-step and M-step until convergence or the maximum number of iterations (20 in our experiments) is achieved, and then obtain the resultant epitome $\hat{\ve}$ from the reference image $cI$.

Note that the above training process is applicable for a single type of feature of $cI$. We use two types of feature to train the epitome, i.e. the YIQ hannels and the dense sift feature \cite{Lazebnik06}. We convert $cI$ from the RGB color space to the YIQ color space where Y channel represents the luminance and IQ channels represent chrominance information. Moreover, dense sift feature is computed for each sampled patch. A $K \times K$ patch is evenly divided into $R \times R$ grids, and the orientation histogram of the gradients with 8 bins is calculate for each grid, which results in a $8R^2$-dimensional dense sift feature vector for each patch. $R$ is typically set as 3 or 4. We then train the epitome $\ve = \left( {{\ve^{YIQ}},{\ve^{dsift}}} \right)$ for the YIQ channels and the dense sift feature, and the epitome for YIQ channels ($\ve^{YIQ}$) share the same hidden mapping with the epitome for the dense sift feature ($\ve^{dsift}$) in the inference process \cite{Ni09}:

\begin{equation} \label{eq:multiepitome}
p(\vZ_k | \vT_k, \ve) =  p(\vZ^{YIQ}_k | \vT_k, \ve^{YIQ})^{\lambda}p(\vZ^{dsift}_k | \vT_k, \ve^{dsift})^{1-\lambda}
\end{equation}

\noindent where $\vZ^{YIQ}_k$ and $\vZ^{sift}_k$ represent the YIQ channel and the dense sift feature of patch $\vZ_k$ respectively, $\ve^{YIQ}$ and $\ve^{dsift}$ represent the epitome trained from the YIQ channels and dense sift feature of $cI$ respectively. $0 \le \lambda  \le 1$ is a parameter balancing the preference between color and dense sift feature.

\subsection{Colorization by Epitome}

With the epitome $\hat\ve$ learnt from the reference image, we colorize the target grayscale image $gI$ by inference in the epitome graphical model. Similar to the epitome training process, we densely sample ${\hat{Q}}$ patches $\{\hat\vZ_k\}^{\hat{Q}}_{k=1}$ from $gI$ (these patches cover the entire $gI$). With the hidden mapping associated with patch $\hat\vZ_k$ denoted as $\hat\vT_k$, the most probable mapping of the patch $\hat\vZ_k$ , i.e. $\hat\vT^*_{k}$, is formulated as below:

\begin{equation} \label{eq:epitomeinfer}
\hat\vT^*_{k} = \mathop {\arg \max }\limits_{\hat\vT_k} p\left( {\hat\vT_k|{\hat\vZ_k},\hat\ve,\vpi } \right)
\end{equation}

\noindent which is essentially the same as the E-step (\ref{eq:estep}). We take the grayscale channel of $gI$ as the luminance channel (Y channel) of itself. Since the color information (IQ channels) is absent in $gI$, we only use the epitomes corresponding to the Y channel and the dense sift feature to evaluate the right hand side of (\ref{eq:epitomeinfer}). The color information is then transferred from the epitome patch, whose location is specified by $\hat\vT^*_{k}$, to the grayscale patch $\hat\vZ_k$. We denote the target image after colorization as $gI_c$. Since $\{\hat\vZ_k\}^{\hat{Q}}_{k=1}$ can be overlapping with each other, the final color (the value of IQ channels) of a pixel $i$ in image $gI_c$ is averaged according to:

\begin{equation} \label{eq:epitomeinfer}
gI_c\left( i \right) = \frac{{\sum\limits_{k = 1}^{\hat{Q}} {\sum\limits_{j \in {\hat{S}_k}} {\delta \left( {j = i} \right)} \hat\ve_{\hat\vT^*_{k}\left( j \right)}^{IQ}} }}{{\sum\limits_{k = 1}^{\hat{Q}} {\sum\limits_{j \in {\hat{S}_k}} {\delta \left( {j = i} \right)} } }}
\end{equation}

where $\hat{S}_k$ is the image coordinates of patch $\hat\vZ_k$, and ${\ve_{\hat\vT^*_{k}\left( j \right)}^{IQ}}$ represents the value of the IQ channels in the epitome $\ve$ at location ${\hat\vT^*_{k}\left( j \right)}$.

\section{Experimental Results}

We show colorization results in this section. As mentioned in section 2, we use square patches of size $K \times K$, and the size of epitome is half of the size of the reference image. We densely sample patches with horizontal and vertical gap of $\omega K$ pixels, where $\omega$ is a parameter between $\left[ {0,1} \right]$ and it controls the number of sampled patches.

\begin{figure*}[!htb]
\begin{center}
\includegraphics[scale=0.6]{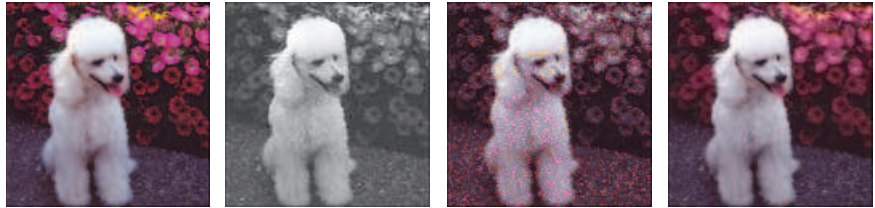}
\end{center}
   \caption{The result of colorizing the dog. From left to right: the reference image, the target image (obtained by converting the reference image to the grayscale), the result by \cite{WelshAM02}, and our result.}
\label{fig:dog}
\end{figure*}

Figure~\ref{fig:dog} shows the result of colorization for the dog image. We convert the original image to grayscale as the target image. The patch size is $12 \times 12$ and the parameter $\lambda$ balancing between the color and the dense sift feature is 0.5. We compare our method to \cite{WelshAM02} which transfers color from the reference image to the target image by pixel-level matching. The result produced by \cite{WelshAM02} lacks spatial continuity and we observe small artifacts throughout the whole image. On the contrary, our method renders a colorized image very similar to the ground truth. This example also demonstrate that the learnt epitome, which is a summary of a large number of sampled patches, contains sufficient color information for colorization.

\begin{figure*}[!htb]
\begin{center}
\includegraphics[scale=0.6]{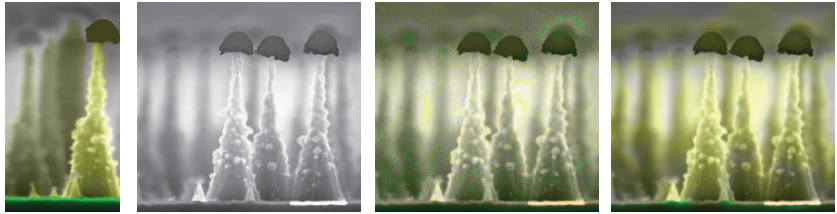}
\end{center}
   \caption{The result of colorizing the Nano Mushroom-like images}
\label{fig:nano}
\end{figure*}

\begin{figure*}[!htb]
\begin{center}
\includegraphics[scale=0.6]{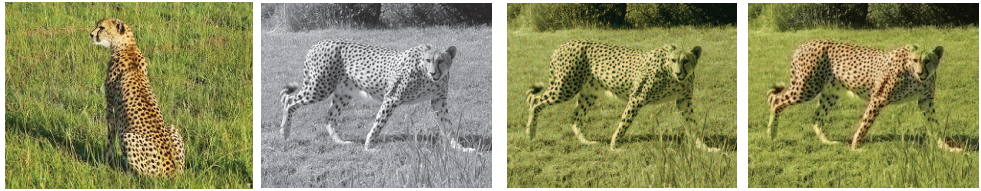}
\end{center}
   \caption{The result of colorizing the cheetah}
\label{fig:cheetah}
\end{figure*}

Figure~\ref{fig:nano} and ~\ref{fig:cheetah} shows the colorization result for the Nano Mushroom-like images and the cheetah. The patch size is chosen as $12 \times 12$ and $15 \times 15$ respectively, and $\lambda$ is set to be 0.8 for both cases. \cite{WelshAM02} still generates artifacts around the top and bottom of the Mushroom-like structure, while our method produce a much more spatially coherent result. Moreover, we transfer the correct color for the cheetah to the target image, which results in a more natural colorization result than that of \cite{WelshAM02}.

\section{Conclusion}

We present an automatic colorization method using epitome in this paper. While most of existing colorization methods require tedious and time consuming user intervention for scribbles or segmentation, our epitomic colorization method is automatic. Epitomic colorization exploits the color redundancy by summarizing the color information in the reference image into a condensed image shape and appearance representation. Experimental results shows the effectiveness of our method.

{
\bibliographystyle{ieee}
\bibliography{epitome}
}

\end{document}